\documentclass[journal]{IEEEtran}

\ifCLASSINFOpdf
\else
\fi
\usepackage{caption}
\usepackage{amsmath}
\usepackage{enumerate}%\begin{enumerate}[i] % reacts to 1, a, A, i, and I
\usepackage{mathrsfs}
\usepackage{makeidx}         % allows index generation
\usepackage{graphicx}        % standard LaTeX graphics tool
\usepackage{multicol}        % used for the two-column index
\usepackage{subfigure}
\usepackage{epsfig,amssymb,latexsym}
\usepackage{psfrag}
\usepackage{algorithm}
\usepackage{algorithmic}
\usepackage{xcolor}
\usepackage{nomencl}

\newtheorem{assumption}{Assumption}%[section]
\newtheorem{remark}{Remark}%[section]
\newcommand{\be}{\begin{equation}}
\newcommand{\ee}{\end{equation}}
\newcommand{\bea}{\begin{eqnarray}}
\newcommand{\eea}{\end{eqnarray}}
\newcommand{\ba}{\begin{array}}
\newcommand{\ea}{\end{array}}
\newcommand{\bc}{\begin{center}}
\newcommand{\ec}{\end{center}}

\usepackage{amsmath}
\DeclareMathOperator*{\argmin}{arg\,min}
\DeclareMathOperator*{\argmax}{arg\,max}

\begin{document}
%
% paper title
% can use linebreaks \\ within to get better formatting as desired
% Do not put math or special symbols in the title.
\title{Role Playing Learning for Socially Concomitant Mobile Robot Navigation}
%
%
% author names and IEEE memberships
% note positions of commas and nonbreaking spaces ( ~ ) LaTeX will not break
% a structure at a ~ so this keeps an author's name from being broken across
% two lines.
% use \thanks{} to gain access to the first footnote area
% a separate \thanks must be used for each paragraph as LaTeX2e's \thanks
% was not built to handle multiple paragraphs
%

\author{Mingming~Li,
        Rui~Jiang,
        Shuzhi~Sam~Ge,~\IEEEmembership{Fellow,~IEEE,}
        and~Tong~Heng~Lee,~\IEEEmembership{Member,~IEEE}% <-this % stops a space

\thanks{M. Li, R. Jiang, S. S. Ge and T. H. Lee are with the Department of Electrical
and Computer Engineering, and the Social Robotics Lab, Smart System Institute (SSI), National University of Singapore,
Singapore 117576 (e-mail: li\_mingming@u.nus.edu; rui\_jiang@u.nus.edu; samge@nus.edu.sg;
eleleeth@nus.edu.sg).}% <-this % stops a space
}

\maketitle

% As a general rule, do not put math, special symbols or citations
% in the abstract or keywords.
\begin{abstract}
In this paper, we present the Role Playing Learning (RPL) scheme for a mobile robot to navigate socially with its human companion in populated environments. Neural networks (NN) are constructed to parameterize a stochastic policy that directly maps sensory data collected by the robot to its velocity outputs, while respecting a set of social norms. An efficient simulative learning environment is built with maps and pedestrians trajectories collected from a number of real-world crowd data sets. In each learning iteration, a robot equipped with the NN policy is created virtually in the learning environment to play itself as a companied pedestrian and navigate towards a goal in a socially concomitant manner. Thus, we call this process Role Playing Learning, which is formulated under a reinforcement learning (RL) framework. The NN policy is optimized end-to-end using Trust Region Policy Optimization (TRPO), with consideration of the imperfectness of robot's sensor measurements. Simulative and experimental results are provided to demonstrate the efficacy and superiority of our method.
\end{abstract}
l
% Note that keywords are not normally used for peerreview papers.

% For peer review papers, you can put extra information on the cover
% page as needed:
% \ifCLASSOPTIONpeerreview
% \begin{center} \bfseries EDICS Category: 3-BBND \end{center}
% \fi
%
% For peerreview papers, this IEEEtran command inserts a page break and
% creates the second title. It will be ignored for other modes.
\IEEEpeerreviewmaketitle

\section{Introduction}
\label{Intro}

The capability to navigate in densely populated and dynamic environments is one of the most important features that enable the deployment of mobile robots in unstructured environment, such as schools, shopping malls and transportation hubs. The key difference between the problem of navigating among humans and the traditional path planning and obstacle avoidance problems is that humans tend to smoothly evade each other interactively and cooperatively, rather than remaining static or maintaining an indifferent trajectory dynamics. In other words, there are social norms that need to be understood and complied to achieve maximum comfort of all involved pedestrians during navigation. We refer to this as the problem of social navigation, which aims to model such social norms and develop a robotic navigation policy that is socially acceptable to the pedestrians around.

For social navigation, the traditional approaches based on Dynamic Window Approach (DWA) \cite{thrun1997dynamic} or potential fields \cite{hwang1992potential,ge2000new} are usually of limited efficacy as pedestrians are simply regarded as uncooperative obstacles. An illustrative example is the freezing robot problem (FRP) \cite{trautman2010unfreezing,trautman2015robot}, where a mobile robot will be stuck in a narrow corridor when facing a crowd of people if it lacks the ability to predict the joint collision avoidance behaviors of human pedestrians. To this end, researches have been done to understand the principles of humans' joint collision avoidance strategies and one of the pioneering works are the social force model (SFM) \cite{helbing1995social,helbing2000simulating}. Other joint collision avoidance model such as reciprocal velocity obstacles (RVO) have been proposed in \cite{van2011reciprocal,van2011lqg,van2008reciprocal}, with an underlying assumption that all involved agents adopt the same collision avoidance strategies. These ideas are also applied to visual tracking of pedestrians \cite{pellegrini2009you,yamaguchi2011you}. More recently, several attempts are made to learn probabilistic models of pedestrians' trajectories during joint collision avoidance, based on which the robot's navigation decision is generated such that it is able to behave naturally and correctly in similar situations \cite{kuderer2012feature,trautman2015robot,kretzschmar2016socially,kim2016socially}.

In this paper, we propose to augment the dimensions of human-robot interaction in social navigation by further endowing robot with appropriate group behaviors when it is travelling with a human companion. This capability is highly desirable for assistive mobile robots \cite{bicchi2010towards,gross2011progress,wang2014adaptive}, which serve as assistants and companions and are expected to travel along with theirx human partners in not only home environment but also possibly crowded public areas. In other words, apart from understanding the collision avoidance behaviors of pedestrians, the robot also needs to consider the motion of its companion so as to maintain a sense of affinity when they are travelling together towards a certain goal. We call this socially concomitant navigation (SCN) and it is more challenging than the aforementioned social navigation problem, where the robot is assumed to travel alone with a simpler pursuit of reaching a specific goal while being free of collision.

To address the problem of SCN, we develop a new learning scheme called Role Playing Learning (RPL). Particularly, we formulate such problem under the framework of Partially Observable Markov Decision Process (POMDP) and reinforcement learning (RL). A neural network (NN) is used to parameterize the navigation policy of the robot, which is optimized to gives proper steering commands for the next time instance based on the robot's current and previous observations to its surroundings. To facilitate the RL process, we create a simulative navigation environment by mirroring a collections of real world pedestrians data sets and develop an on-policy optimization method called Partially Observable Trust Region Policy Optimization (PO-TRPO). In each run in an optimization iteration, the robot will attempt to play itself as a companion of a randomly chosen pedestrian by executing the NN navigation policy. The NN policy is then optimized using PO-TRPO based on a batch of collected trajectories. Compared to the existing analytically derived or data-driven approaches, our RPL scheme has the following advantages:

\begin{enumerate}
  \item RPL scheme is less restrictive. It does not rely on the assumption that the robot and other agents (pedestrians) share the same decision-making models \cite{van2011reciprocal,van2011lqg,van2008reciprocal,trautman2015robot,kretzschmar2016socially} or that the navigation goals of pedestrians are known \cite{trautman2015robot,kretzschmar2016socially}.
  \item The formulation of RPL scheme is more generalizable and flexible. Our formulation contain no manually-defined feature and domain knowledge (e.g., statistics of pedestrians' behaviors). It is not hardware-specific and can be easily modified to incorporate kinematics of different mobile robot platforms, sensor specifications and navigation objectives. In addition, unlike \cite{kim2016socially,kretzschmar2016socially}, the learned navigation policy operates without assess to the global map of the environment. Therefore, it is not environment-specific and is well generalizable to unmet real-world scenarios.
  \item We explicitly consider the noise and limitation of the robot's sensor measurements. Most approaches for social navigation assume that the robot has full and accurate knowledge of interested variables, such as positions or distance of pedestrians and obstacles \cite{van2011reciprocal,van2011lqg,van2008reciprocal,kretzschmar2016socially}. On the contrary, our RPL schemes is rooted from the situation where the robot can only perceive those lie within its sensor's Field of View (FoV), with the existence of measurement noise.
  \item As a RL-based approach, RPL is efficient. Although RPL aims at solving tasks that involve interaction among robot, humans and physical environment, it does not require participation of human in both data collection and learning, which is known to be tedious and time-consuming. Instead, the learning process is safely automated in a simulative yet realistic environment with no human intervention.
\end{enumerate}

We evaluate the performance of our approach in both simulations and real-world experiments, by comparing it with a baseline planner based on RVO \cite{van2011reciprocal} and humans, repectively. We also show that, with some tricks, the learned navigation policy can still be effective when the navigation scenario is reduced to the aforementioned social navigation, which means the robot is travelling without human companion.

The remainder of this paper is organized as follows. Related work is first reviewed in Section II. In Section III, the problem of SCN is formulated as a POMDP and associated definitions are given. RPL scheme and PO-TRPO algorithm are described in Section IV. Sections V and VI provide extensive results of simulation and experiment, followed by some concluding remarks in Section VII.

\section{Related Work}
The problems of robot navigation in populated and dynamic environment can be addressed from a number of angles, which can be largely classified into two groups as in the following subsections.

\subsection{Interactive Behaviors Models}
Many researches have been proposed to describe the interactive navigation behaviors of humans by fitting a computational model to the observed pedestrians trajectories \cite{argall2009survey}. In this way, the robot's path planner is able to understand pedestrians' intention during joint collision avoidance and actively calculate an optimal route towards its goal.

In the field of robotics, a majority of work in this direction is done via inverse reinforcement learning (IRL) \cite{abbeel2004apprenticeship}, which learns a cost function that explains the observed behaviors. For example, maximum entropy IRL \cite{ziebart2008maximum} is adopted in a number works \cite{ratliff2006maximum,ziebart2009planning,henry2010learning,vernaza2012efficient,kitani2012activity} for discrete human behavior prediction and route planning. However, discrete representation is less desirable when modeling trajectories, which are in nature continuous and has higher order dynamics, such as velocities and acceleration. Instead, \cite{kim2016socially} adopts Maximum-A-Posteriori Bayesian IRL \cite{choi2011map} to learn appropriate navigation behavior of a specific mobile robot from a set of demonstration trajectories. Note that, the demonstration data in \cite{kim2016socially} is specific to configurations of the robot and its sensor and has to be collected via human operation, which could be time-consuming. On the other hand, \cite{kuderer2012feature,kretzschmar2016socially} learns probabilistic models of composite trajectories of pedestrians from video data by maximum entropy learning and IRL. To better capture the characteristics of observed trajectories, they propose to develop their models based on a set of features that are hand-crafted according to the domain knowledge from psychological studies. In addition, those features contain velocities and accelerations of pedestrians, which, in practice, are hard to precisely measure. Besides, interacting Gaussian process (IGP) is derived in \cite{trautman2015robot} to model the joint trajectories of pedestrian while explicitly considering the effects of observation noise. Nevertheless, the design of IGP also requires several hand-crafted kernels that are formulated based on the priori information in a specific application scenario.

Other than researchers in robotics, the community of computer vision also possess great interest in pedestrian modeling. One of the important topics is trajectory prediction in video space. In \cite{pellegrini2009you}, Linear Trajectory Avoidance (LTA) is developed as a dynamic model for pedestrians in video space for short-term trajectory prediction and it is integrated into visual tracking system. Gaussian process is adopted in \cite{kim2011gaussian} to learn the motion pattern of pedestrians. Recently, Social LSTM is proposed in \cite{alahi2016social} for human trajectory prediction in crowd space. Similarly, the feature of social sensitivity is developed in \cite{robicquet2016learning} to analyze trajectories of pedestrians and bicyclists. While the above methods can effectively predict the navigation intention of pedestrians in videos, it is still unclear how to apply these model to navigation of robot in real scenarios.

\subsection{Steering Models}
In contrast to learning behavior models of pedestrians, a more direct perspective is to develop a steering model that outputs the immediate navigation actions given the robot's current observation to the environment. One of the pioneering work in this direction is the social force model (SFM) \cite{helbing1995social}, which uses energy/potential functions to encode the social status of pedestrian. Then, the navigation motivation of a pedestrian can be derived by taking the gradients of these energy functions. Following this idea, subsequent work \cite{johansson2007specification,lerner2007crowds,helbing2011pedestrian,yamaguchi2011you} propose to infer the optimal parameters of the energy function by fitting them to video data. However, they are likely to produce suboptimal results if the demonstration data from humans are imperfect. In \cite{muller2008socially}, the authors integrate a people tracker and an iterative $\text{A}^*$ planner, with which the robot actively follows the pedestrian travelling in a similar direction to navigate through crowded environment. \cite{mehta2016autonomous} follow the same idea and formulate the choice of a pedestrian to follow as a Multi-Policy Decision Making process. On the other hands, \cite{foka2010probabilistic} develops a hierarchical POMDP for predictive navigation in dynamic environment. The idea is to predict the motion of pedestrians and generate a environment-specific cost map for path planning and obstacle avoidance.

Other than navigating in a pedestrian-aware manner, several reactive collision avoidance techniques have also been developed, such as DWA \cite{thrun1997dynamic,seder2007dynamic}, velocity obstacles \cite{fiorini1998motion} and reciprocal velocity obstacles (RVO) \cite{van2011reciprocal,van2011lqg,van2008reciprocal}. The common idea of these methods is to treat pedestrians as moving obstacles and reactively update the planner every short periods to achieve collision avoidance. As mentioned in Section I, these methods are less effective for social navigation as they lack predictive abilities and are based on some restrictive assumptions, such as accurate knowledge of moving agents' velocities \cite{seder2007dynamic} and that all agents adopt the identical collision avoidance strategy \cite{van2011reciprocal,van2011lqg,van2008reciprocal}.

Our proposed navigation policy belongs to the steering models. It takes an observation vector as input and outputs the navigation action through a stochastic neural networks. During RPL, our policy is optimized by the PO-TRPO algorithm, which is derived based on the recent advances in deep reinforcement learning (DRL) \cite{schulman2015high,schulman2015trust}. DRL exploits the massive representation power of deep neural networks (DNN) \cite{lecun2015deep} to build a complex yet sophisticated decision model, with which an agent can directly learn from raw signals instead of carefully crafted feature and tends to act more intelligently. Recently, there are several attempts in using DNN and DRL for robot navigation. For example, an end-to-end motion planner is learned in \cite{pfeiffer2016perception} to map raw sensor data of a laser range finder onto steering commands of a mobile robot. In \cite{chen2016decentralized}, a decentralized multi-agent collision avoidance policy is learned via DRL, which can be thought as a DRL version of the original RVO approach \cite{van2011reciprocal}. Finally, a target-driven visual navigation policy for home environment is learned in \cite{zhu2016target} via DRL. They create a set 3D virtual home environments for effective and efficient training of the agent, which shares a similar idea with our proposed RPL scheme.

\section{Problem Formulation}
To formulate the problem of socially concomitant navigation, we gives the following rules of SCN:
\begin{enumerate}
  \item The robot should reach its goal as fast as possible;
  \item The robot should not collide with any of the pedestrians or its companion, or run into any obstacle;
  \item The robot should not run too far away from its companion.
\end{enumerate}

The above rules serve as a generic description of the robot's desired performance during navigation. To give concrete definitions, consider the navigation process as an infinite-horizon discounted POMDP in discrete time, defined by the tuple $(\mathcal{S,A},F,\mathcal{O},p_{0},r,\gamma)$. $\mathcal{S}$ is a finite set of states $s$ reflecting the navigation status of the robot. $\mathcal{A}$ is a finite set of actions $a$. In this paper, it is defined as a twosome of the translational and rotational velocities of a synchro-drive mobile robot, i.e., $a=[v_{T},v_{R}]$. $F:\mathcal{S}\times\mathcal{A}\rightarrow\mathcal{S}$ is state-transition mapping, which is characterized by the dynamics of the robot, the other humans and the environment. Without loss of generality, we assume deterministic state transition, i.e., $s_{i+1}=F(s_{i},a_{i})$, where $s_{i},a_{i}$ are the state and action taken at time $t_{i}$. $\mathcal{O}$ is the set of the robot's observation $o$ to the state $s$ and $\beta(o|s)$ denotes the conditional observation probability distribution. Note that, in practice, the robot's observation has only incomplete access to $s$ or is subject to certain measurement noise, which implies $o\neq s$.  $p_{0}:\mathcal{S}\rightarrow\mathbb{R}$ is the initial state distribution, i.e., $s_{0}\sim p_{0}$. $r:\mathcal{S}\rightarrow \mathbb{R}$ is a scalar reward given to the robot and $\gamma\in(0,1]$ is the reward discount factor.

\textbf{Robot motion dynamics:} In this paper, synchro-drive mobile robots are considered, whose motion equation can be approximated by assuming the robot's velocities to be constant within a certain short time period $[t_{i},t_{i+1}]$ \cite{thrun1997dynamic} with length $\Delta t=t_{i+1}-t_i$. Particularly, let $\phi_r(t_i)$ and  $\rho_r(t_i)=[x_{r}(t_i),y_{r}(t_i)]$ denote the robot's heading and its positions in a 2D Cartesian space at time $t_i$, respectively. $v_{T}(t_i)\in[0,\bar{v}_{T}]$ and $v_{R}(t_i)\in[-\bar{v}_{R},\bar{v}_R]$ represent the robot's translational and rotational velocities. Define $\Delta x_r=x_r(t_{i+1})-x_r(t_i)$ and $\Delta y_r=y_r(t_{i+1})-y_r(t_i)$. When the robot has nonzero rotational velocity, i.e., $v_R(t_i)\neq 0$, we have
\begin{eqnarray}
% \nonumber to remove numbering (before each equation)
  \Delta x_r \hspace{-2mm}&=&\hspace{-2mm}-\frac{v_T(t_i)(\sin\phi_r(t_i)-\sin(\phi_r(t_i)+v_R(t_i)\Delta t))}{v_R(t_i)}\label{deltaX} \\
  \Delta y_r \hspace{-2mm}&=&\hspace{-2mm} \frac{v_T(t_i)(\cos\phi_r(t_i)-\cos(\phi_r(t_i)+v_R(t_i)\Delta t))}{v_R(t_i)} \label{deltaY0}
\end{eqnarray}
Otherwise, when $v_R(t_i)=0$,
\begin{eqnarray}
  \Delta x_r &=& v_T(t_i)\cos\phi_r(t_i)\label{deltaX0}\\
  \Delta y_r &=& v_T(t_i)\sin\phi_r(t_i)\label{deltaY}
\end{eqnarray}

With the above formulations, our goal is optimizing a stochastic navigation policy $P_{\theta}:\mathcal{O}\times\mathcal{A}\rightarrow[0,1]$ with parameters $\theta$ in order to maximize the expected discounted reward:

\begin{equation}\label{eta}
    \eta(P_\theta)=\mathbb{E}_{\tau}[\sum_{i=0}^{\infty}\gamma^{i}r(s_{i},a_{i})]
\end{equation}
where $\tau=(s_{0},o_{0},a_{0},s_{1},o_{1},a_{1},\cdots)$ denotes the whole trajectory and $a_{i}\sim P_\theta(a_{i}|o_{i})$. The specific definitions of the above ingredients for SCN will be elaborated as follows:

\textbf{State:} Given $\rho_r$ and $\phi_r$, define the distance $d$ and direction $\phi$ of a point $\rho=[x,y]$ to the robot as follows:
\begin{eqnarray}
% \nonumber to remove numbering (before each equation)
  d(\rho) &=& \sqrt{(x-x_r)^2+(y-y_r)^2}\label{d} \\
  \phi(\rho) &=& \arctan(\frac{y-y_r}{x-x_r})-\phi_r\label{phi}
\end{eqnarray}

Then, the robot's distance to the goal located at $\rho_{g}=[x_{g},y_{g}]$ are computed as $d_{g}=d(\rho_{g})$ and $\phi_{g}=\phi(\rho_g)$ denotes the offset angle between the robot's current heading $\phi_{r}$ and its goal. Similarly, we can define the twosomes $(d_{\text{ped}}^{j},\phi_{\text{ped}}^{j})$, $(d_{\text{com}}^{j},\phi_{\text{com}}^{j})$ or $(d_{\text{obs}}^{j},\phi_{\text{obs}}^{j})$ to describe the relative position of a pedestrians $\rho_\text{ped}^{j}$, a companion $\rho_\text{com}^{j}$ or an obstacle $\rho_\text{obs}^{j}$ to the robot. With such definitions, the state $s$ is defined to incorporate the information related to the robot's navigation status as follows:

\begin{equation}\label{state}
    s=[d_{g},\phi_{g},a,p_{\text{ped}},p_{\text{com}},p_{\text{obs}}]
\end{equation}
where $a$ is the current action vector and
\begin{eqnarray}
% \nonumber to remove numbering (before each equation)
  p_{\text{ped}} &\hspace{-3mm}=\hspace{-3mm}& [d_{\text{ped}}^1,\phi_{\text{ped}}^1,\cdots,d_{\text{ped}}^{n_\text{ped}},\phi_{\text{ped}}^{n_\text{ped}}]\label{p} \\
  p_{\text{com}} &\hspace{-3mm}=\hspace{-3mm}& [d_{\text{com}},\phi_{\text{com}}]\label{c}  \\
  p_{\text{obs}} &\hspace{-3mm}=\hspace{-3mm}& [d_{\text{obs}}^{F},d_{\text{obs}}^{L^-},\phi_{\text{obs}}^{L^-},d_{\text{obs}}^{R^-},\phi_{\text{obs}}^{R^-},d_{\text{obs}}^{L^+}\hspace{-0.5mm},\phi_{\text{obs}}^{L^+},d_{\text{obs}}^{R^+}\hspace{-0.5mm},\phi_{\text{obs}}^{R^+}\hspace{-0.5mm}]\label{z}
\end{eqnarray}
The vector $p_{\text{ped}}$ includes the distances and directions of $n_\text{ped}$ closest pedestrians while $p_{\text{com}}$ includes those of the robot's companion.

The vector $p_{\text{obs}}$ is a compact description to the robot's perception of the surrounding environment. Particularly, the boundaries of the occupied space (obstacles) in the environment are represented as a finite point set $\mathbb{Z}=\{\rho_{\text{obs}}^{1},\rho_{\text{obs}}^{2},\cdots,\rho_{\text{obs}}^{j},\cdots\}$. Then, the 9 variables in $p_\text{obs}$ are defined based on the following assumption
{\assumption\label{assump.farObs} An obstacle $\rho_{\text{obs}}\in\mathbb{Z}$ has no effect on the robot's navigation decision if it satisfies $d(\rho_{\text{obs}})>\bar{d}_\text{obs}$, where $\bar{d}_\text{obs}$ is a predefined finite constant.
}

By \textbf{Assumption} \ref{assump.farObs}, it is sufficient to consider only obstacles in $\mathbb{Z}$ that are closed enough to the robot, whose distances are less than $\bar{d}_\text{obs}$. In practice, this limit may correspond to the robot's perception range. Let
\begin{equation}\label{barZ}
  \bar{\mathbb{Z}}=\{\rho|\rho\in\mathbb{Z},d_{\rho}\leq\bar{d}_{\text{obs}}\}
\end{equation}
The components in vector $p_{\text{obs}}$ are described as follows:

The distance to the nearest obstacle located at heading of the robot, i.e.,
\begin{equation}\label{dobsF}
    d_{\text{obs}}^{F}=\min_{\rho\in\bar{\mathbb{Z}}\text{ and } |\phi(\rho)|\leq\epsilon_\rho}d(\rho)
\end{equation}
where $\epsilon_\rho$ is a small constant.

For $d_{\text{obs}}^{L},\phi_{\text{obs}}^{L},d_{\text{obs}}^{R},\phi_{\text{obs}}^{R}$, they represent the distance and direction of the closest and farthest obstacles on the robot's left ($\rho_\text{obs}^L$) and right side ($\rho_\text{obs}^R$), respectively, which are defined mathematically as follows:
\begin{eqnarray}
% \nonumber to remove numbering (before each equation)
  \rho_\text{obs}^{L^-} &=& \argmin_{\rho\in \bar{\mathbb{Z}} \text{ and } \phi(\rho)>\epsilon_\rho}d(\rho)\label{rhoL} \\
  \rho_\text{obs}^{R^-} &=& \argmin_{\rho\in \bar{\mathbb{Z}} \text{ and } \phi(\rho)<-\epsilon_\rho}d(\rho)\label{rhoR}\\
  \rho_\text{obs}^{L^+} &=& \argmax_{\rho\in \bar{\mathbb{Z}} \text{ and } \phi(\rho)>\epsilon_\rho}d(\rho)\label{rhoLMax} \\
  \rho_\text{obs}^{R^+} &=& \argmax_{\rho\in \bar{\mathbb{Z}} \text{ and } \phi(\rho)<-\epsilon_\rho}d(\rho)\label{rhoRMax}
\end{eqnarray}

\begin{figure}
  % Requires \usepackage{graphicx}
  \includegraphics[width=\hsize]{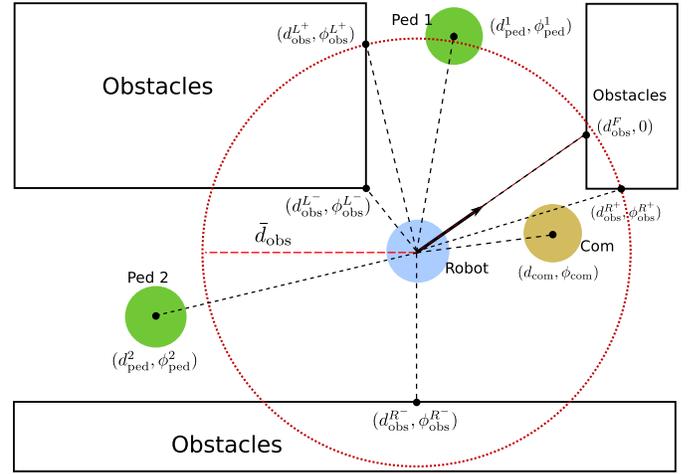}\\
  \caption{Illustration of the state variables in Eq. (\ref{state}). The blue, yellow and green circles represent the robot, its companion (Com.) and the pedestrians (Ped.) respectively. The red dashed circle with a radii $\bar{d}_{\text{obs}}$ represents the boundary of the set $\bar{\mathbb{Z}}$ in Eq. \ref{barZ}. The black arrow shows the current heading of the robot. Considering the robot's current position as the origin, the polar coordinates of the pedestrians, the companion, the closest( and the farthest) obstacles in each direction are compactly represented as vectors $p_{\text{ped}},p_{\text{com}},p_{\text{obs}}$. }\label{fig.state}
\end{figure}

Then, the variables in $p_{\text{obs}}$ can be simply determined as the distance and directions of the above points according to Eqs. (\ref{d}) and (\ref{phi}). Figure. \ref{fig.state} provides a comprehensive illustration of the state variables $p_{\text{ped}},p_{\text{com}}$ and $p_{\text{obs}}$.

\begin{figure}[h]
\centering
  % Requires \usepackage{graphicx}
  \includegraphics[width=0.6\hsize]{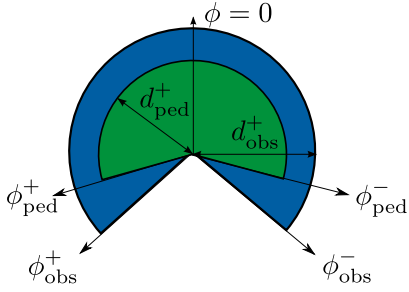}\\
  \caption{Field of Views of the pedestrians (green) and obstacles (blue) detectors. The arrow ($\phi=0$) points towards the current heading of the robot. The constants $\phi_\text{ped}^+,\phi_\text{obs}^+$ and $\phi_\text{ped}^-,\phi_\text{obs}^-$ denote the maximum and minimum offset angles in the corresponding FoVs. Finally, $d_\text{ped}^+$ and $d_\text{obs}^+$ represent the maximum detection ranges for the pedestrian and obstacle detectors, respectively. The values of these constants should be determined according the specific configurations of the robot's sensor and the corresponding detection algorithms. Any pedestrian/obstacle outside the FoVs is not observable and therefore will be omitted.}\label{fig.fov}
\end{figure}

\textbf{Observation:} As discussed in the previous sections, sensors mounted on the robot are always subject to various kinds of limitation and measurement noise, which must be taken into account in order to develop a robust and practical navigation system. To this end, we define $o$ as the robot's observation to the true state $s$ as follows:
\begin{equation}\label{observation}
    o=[d_{g},\phi_{g},a,\hat{p}_{\text{ped}},\hat{p}_{\text{com}},\hat{p}_{\text{obs}}]
\end{equation}
By Eq. (\ref{observation}), we assume that the robot has accurate information about the goal position and its current taken action (i.e., the velocity commands output to the robot's motor) while its observations to $p_{\text{ped}},p_{\text{com}},p_{\text{obs}}$ may be imperfect. Particularly, consider the Field of Views (FoVs) for the robot's pedestrian and obstacle detectors illustrated as Fig. \ref{fig.fov}.

Mathematically, let finite point sets $\mathbb{F}_\text{ped}$ and $\mathbb{F}_\text{obs}$ denote the current FoVs of pedestrian and obstacle detectors, characterized by threesomes $(\phi_\text{ped}^+,\phi_\text{ped}^-,d_\text{ped}^+)$ and $(\phi_\text{obs}^+,\phi_\text{obs}^-,d_\text{obs}^+)$, respectively. The robot's observations to the pedestrians' relative positions are obtained as
\begin{equation}\label{hatpped}
    \hat{p}_\text{ped}=[\hat{d}_{\text{ped}}^1,\hat{\phi}_{\text{ped}}^1,\cdots,\hat{d}_{\text{ped}}^{n_\text{ped}},\hat{\phi}_{\text{ped}}^{n_\text{ped}}]
\end{equation}
where
\begin{eqnarray}\label{hatdp}
\hat{d}_{\text{ped}}^{j}=
  \left\{
  \begin{array}{ll}
  d_{\text{ped}}^{j}+\tilde{d}_\text{ped}, & \hbox{\text{if} $\rho_\text{ped}^j\in\mathbb{F}_\text{ped}$}\\
  d_{\text{ped}}^{+}, & \hbox{else}\\
  \end{array}
  \right.
\end{eqnarray}
and
\begin{eqnarray}\label{hatphip}
\hat{\phi}_{\text{ped}}^j=
  \left\{
  \begin{array}{ll}
  \phi_{\text{ped}}^{j}, & \hbox{\text{if} $\rho_\text{ped}^j\in\mathbb{F}_\text{ped}$}\\
  \pi, & \hbox{\text{if} else}\\
  \end{array}
  \right.
\end{eqnarray}
for $j=1,\cdots,n_\text{ped}$, with $\tilde{d}_\text{ped}$ being the measurement noise/error.

Similarly, define
\begin{equation}\label{hatpobs}
    \hat{p}_{\text{obs}}=[\hat{d}_{\text{obs}}^{F},\hat{d}_{\text{obs}}^{L^-},\hat{\phi}_{\text{obs}}^{L^-},\hat{d}_{\text{obs}}^{R^-},
    \hat{\phi}_{\text{obs}}^{R^-},\hat{d}_{\text{obs}}^{L^+},\hat{\phi}_{\text{obs}}^{L^+},\hat{d}_{\text{obs}}^{R^+},\hat{\phi}_{\text{obs}}^{R^+}]
\end{equation}
Compared to the states in (\ref{dobsF}) to (\ref{rhoRMax}), only the obstacles within $\mathbb{F}_\text{obs}$ are observable. Thus, $\hat{d}_\text{obs}^F$ is formulated as
\begin{equation}\label{hatdobsF}
% \nonumber to remove numbering (before each equation)
  \hat{d}_\text{obs}^F = \min_{\rho\in \bar{\mathbb{Z}}\cap\mathbb{F}_\text{obs} \text{ and } |\phi(\rho)|\leq\epsilon_\rho}d(\rho)+\tilde{d}_{\text{obs}}
\end{equation}
where $\tilde{d}_{\text{obs}}$ is the measurement noise/error for obstacle detection. The closest observed obstacles on the robot's left and right sides are defined in a similar way as:
\begin{eqnarray}
% \nonumber to remove numbering (before each equation)
  \hat{\rho}_\text{obs}^{L^-} &=& \argmin_{\rho\in \bar{\mathbb{Z}}\cap\mathbb{F}_\text{obs} \text{ and } \phi(\rho)>\epsilon_{\rho}}d(\rho)+\tilde{d}_{\text{obs}}\label{hatRhoL} \\
  \hat{\rho}_\text{obs}^{R^-} &=& \argmin_{\rho\in \bar{\mathbb{Z}}\cap\mathbb{F}_\text{obs} \text{ and } \phi(\rho)<-\epsilon_{\rho}}d(\rho)+\tilde{d}_{\text{obs}}\label{hatRhoR}\\
    \hat{\rho}_\text{obs}^{L^+} &=& \argmax_{\rho\in \bar{\mathbb{Z}}\cap\mathbb{F}_\text{obs} \text{ and } \phi(\rho)>\epsilon_{\rho}}d(\rho)+\tilde{d}_{\text{obs}}\label{hatRhoLMax} \\
  \hat{\rho}_\text{obs}^{R^+} &=& \argmax_{\rho\in \bar{\mathbb{Z}}\cap\mathbb{F}_\text{obs} \text{ and } \phi(\rho)<-\epsilon_{\rho}}d(\rho)+\tilde{d}_{\text{obs}}\label{hatRhoRMax}
\end{eqnarray}
Then, their distance and directions to the robot are calculated using Eqs. (\ref{d}) and (\ref{phi}). For observation to the robot's companions, we rely on the following assumptions.
{\assumption The companions $\rho_\text{com}^1,\cdots,\rho_\text{com}^{n_\text{com}}$ are always observable to the robot.
}

Then, $\hat{p}_{\text{com}}=[\hat{d}_{\text{com}},\phi_{\text{com}}]$, where
\begin{equation}\label{hatdc}
\hat{d}_{\text{com}}=d_{\text{com}}+\tilde{d}_{\text{com}}
\end{equation}

{\remark By Eqs. (\ref{hatdp}) and (\ref{hatdobsF}) to (\ref{hatdc}), it is implied that the observation/measurement noises $\tilde{d}_\text{ped},\tilde{d}_\text{obs}$ and $\tilde{d}_\text{com}$ are additive and independent in different observations. A typical example of such noise is the Additive Gaussian White Noise (AGWN).
}

{\remark Our general formulations of states (\ref{state}) and observations (\ref{observation}) are applicable to various types of onboard sensors, such as range sensors \cite{choi2016extrinsic,miller2015optimal}, RGB-D \cite{endres20143}, Time-of-Flight (ToF) \cite{foix2010object} and omnidirectional cameras \cite{liu2014topological}, as long as the interested positions can be extracted/estimated from the sensor's raw measurements.
}

{\remark The mathematical definitions of the variables in observations $\hat{p}_{\text{ped}},\hat{p}_{\text{com}},\hat{p}_{\text{obs}}$ are given for better understanding and are required only in the simulative RPL process. In practice, it is clear that these values can be directly measured via the robot's onboard sensors without accessing the actual 2-D Cartesian coordinates $[x,y]$ of the considered point sets (e.g., $\mathbb{Z}$, $\mathbb{F}_\text{ped}$ and $\mathbb{F}_\text{obs}$). For example, consider a robot equipped with a laser range finder. These distances and offset angles can be easily obtained from the returned ranges array\cite{kneip2009characterization}.
}

\textbf{Reward function:} A scalar reward will be given to the robot as an award of reaching the goal or a penalty of colliding with obstacles/pedestrians/companions or losing its companions. Particularly, at time $t_i$, the process of SCN will be terminated if any of the following three termination conditions is true.

1) Goal Reaching Condition
\begin{equation}\label{GoalReaching}
  d_{g}(t_i)\leq0.8
\end{equation}

2) Collision Conditions
\begin{eqnarray}
% \nonumber to remove numbering (before each equation)
  \min_{j}d_{\text{ped}}^j(t_i)&\leq& 0.4 \label{PedCollision}\\
  d_{\text{com}}(t_i)&\leq& 0.4 \label{ComCollision}\\
  \min(d_{\text{obs}}^F(t_i),d_{\text{obs}}^{L^-}(t_i),d_{\text{obs}}^{R^-}(t_i))&\leq& 0.2\label{ObsCollision}
\end{eqnarray}

3) Stray Condition
\begin{equation}\label{StrayCondition}
  d_{\text{com}}(t_i)\geq 2
\end{equation}

%Otherwise, a continuous reward $r(t_{i})$ dependent on the current states $s_i$ will be assigned to the robot based on the following penalty function:
%\begin{equation}\label{penalty}
%\small
%  \kappa(s_i)\hspace{-1mm}=\hspace{-1mm}-5d_{g}(t_i) + 5\min(\min_{j}d_{\text{ped}}^j(t_i),2) - |d_{\text{com}}-d_{\text{com}}^*| + \min(d_\text{obs}^F,2)
%\end{equation}
Based on the above three terminal conditions, a reward $r$ will be given to the robot as follows:
\begin{equation}\label{r}
r=
  \left\{
  \begin{array}{llll}
  10000, &\hbox{\text{if} (\ref{GoalReaching})}\\
  -10000, &\hbox{\text{if} (\ref{PedCollision}) \text{or} (\ref{ComCollision}) \text{or}}\\
  &\hbox{~~~~((\ref{ObsCollision}) or (\ref{StrayCondition})}\\
  -10|v_{R}|, &\hbox{else}\\
  \end{array}
  \right.
\end{equation}

Clearly, a positive reward will be given to the robot if it reaches its goal and it will receive a large negative reward if it collides with anything or be stray from its companion. Otherwise, the robot will receive an intermediate reward $-10|v_{R}|$, which penalizes the robot for its rotational velocity to encourage a smoother trajectory with less turning behaviors.

\section{Role Playing Learning}
In this section, we described the RPL scheme to learn an effective navigation policy $P_{\theta}(a|o)$ for SCN in an efficient data-driven manner. The core idea is to transform the crowd trajectories data collected from real-world into a simulative and dynamic navigation environment, where the robot can play itself as a virtual pedestrian and iteratively improve the performance of $P_{\theta}(a|o)$ via Partially Observable Trusted Region Policy Optimization (PO-TRPO).

Consider a set of simulative navigation environment $\mathbb{E}=\{\mathcal{E}_{1},\cdots,\mathcal{E}_j,\cdots\}$. Each environment $\mathcal{E}_{j}=(\mathbb{T}_j,\mathcal{M}_j)$ contains a set of pedestrian trajectories $\mathbb{T}_{j}=\{\rho_{0:T_{k}}^k\}$ and a binary map $\mathcal{M}_j$ that annotates the 2-D Cartesian coordinates of obstacles/occupied space in the environment. With $\mathbb{E}$, the abstract process of RPL is described by the following pseudo codes in \textbf{Algorithm} \ref{RPL}.
\begin{algorithm}[h]
\caption{Role Playing Learning}
\begin{algorithmic}\label{RPL}
\STATE Initialize navigation policy $P_{\theta}$
\FOR{$\text{Iter}=0,1,\cdots,\text{MaxIter}$}
\WHILE {Number of collected sample time steps $\leq$ Batch\_size}
\STATE Randomly choose an environment $\mathcal{E}_j$ from $\mathbb{E}$ and then a trajectory $\rho_{0:T_k}^k$ from $\mathbb{T}_j$.
\STATE Initialize the robot's position at $\rho_{0}^k$ and initial velocities $[v_{T},v_{R}]=[0,0]$. Set $\rho_g=\rho_{T_k}^k$. Choose the robot's heading such that $\phi_g =0$.
\STATE Choose SCN Mode with probability 0.5
\IF{SCN Mode}
\STATE Assign $\rho_{T_{0}:T_k}^k$ as the trajectory of the robot's companion, where $T_{0}=\argmin_{T'}\|\rho_{T'}^k-\rho_{T_k}^k\|\geq 0.6$
\ELSE
\STATE Create a synthesized companion that moves along the robot
\ENDIF
\STATE Assign all other trajectories in $\mathbb{T}_j$ as pedestrians.
\WHILE {None of the termination conditions in (\ref{GoalReaching}) to (\ref{StrayCondition}) is satisfied}
\STATE Update the states and observations of the robot according to Eqs. (\ref{state}) and (\ref{observation}).
\STATE Let the robot execute its policy $P_{\theta}$.
\STATE Update the robot's position according to dynamics (\ref{deltaX}) to (\ref{deltaY})
\STATE Calculate the current reward from Eq. (\ref{r})
\STATE Update the positions of the companion and pedestrians according to the trajectories in $\mathbb{T}_{j}$.
\ENDWHILE
\ENDWHILE
\STATE Update $P_\theta$ using PO-TRPO.
\ENDFOR
\end{algorithmic}
\end{algorithm}

\textbf{Companion Synthesization in non-SCN mode:} As described in \textbf{Algorithm} \ref{RPL}, RPL actually incorporates two different navigation scenarios: the SCN proposed in this paper and the traditional social navigation scenario, where the robot has no human companion. This helps develop a navigation policy adaptable to both situations, with no restrictive assumption on the existence of companion. Particularly, the companion position vector $p_{\text{com}}$ and its observation $\hat{p}_{\text{com}}$ are synthesized, with $d_{\text{com}}=\hat{d}_{\text{com}}=0.8$ for every time step while $\phi_{\text{com}}=\phi_g$. It is clear that the synthesized $p_{\text{com}}$ is equivalent to the situation where the companion is travelling non-distractively along the robot with a constant distance and guarantee that termination conditions (\ref{ComCollision}) and (\ref{StrayCondition}) are always false.

On the other hand, in SCN mode, the companion is assigned with a truncated trajectory $\rho_{T_{0}:T_k}^k$ such that the initial robot-companion distance is sufficiently large.

\begin{figure}
\centering
  % Requires \usepackage{graphicx}
  \includegraphics[width=\hsize]{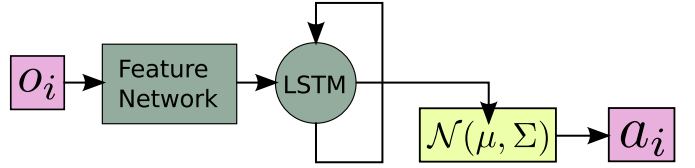}\\
  \caption{Structure of the deep policy network $P_{\theta}$. At time $t_{i}$, the observation vector $o_{i}$ is input to the feature network, which is a feedforward multi-layer perceptron (MLP). The output of the feature network is then fed to a LSTM network \cite{hochreiter1997long}, a recurrent network for aggregation of the information collected through the navigation process. The LSTM network's outputs are assigned as the mean vector $\mu\in\mathbb{R}^2$ of the diagonal Gaussian unit $\mathcal{N}(\mu,\Sigma)$ on the right. The covariance matrix $\Sigma=\sigma^2 I\in\mathbb{R}^{2\times2}$, however, is independent of $o_{i}$ amd it is designed to be gradually decreasing during training and fixed during tests and experiments. Finally, the actions $a_{i}=[v_{T},V_{R}]$ are drawn according to $\mathcal{N}(\mu,\Sigma)$.}\label{network}
\end{figure}
In this paper, we construct a deep policy neural network to parameterize the navigation policy $P_\theta$, whose structure is shown in Fig. \ref{network}. The policy network $P_{\theta}$ is to be trained with the Trust Region Policy Optimization (TRPO) \cite{schulman2015trust} method. However, the original TRPO method is derived based on fully observable MDP, which can not be directly applied to our problem due to the imperfect observation in our formulation and practice. Thus, we proposed to extend the original TRPO algorithm as PO-TRPO, which will be described in the following subsections.

\subsection{Trusted Region Policy Optimization}
The TRPO \cite{schulman2015trust} algorithm is an effective on-policy optimization method for large nonlinear policies and tends to give monotonic improvement during the iterative optimization process. To be specific, a fully observable MDP is considered by TRPO and therefore the policy to be optimized is formulated as $P^*_{\xi}(a|s)$, where $\xi$ is the parameter vector of the policy $P^*$. Note that, $P^*_{\xi}(a|s)$ determines the action $a$ directly from the true state $s$, which differs from our observation-based policy $P_{\theta}(a|o)$. Let us consider the following standard definitions of the state-action value function $Q_{\xi}(s_{i},a_i)$, the value function $V_{\xi}(s_{i})$ and the advantage function $A_{\xi}(s_{i},a_i)$:
\begin{equation}\label{Q}
    Q_{\xi}(s_{i},a_i)=\mathbb{E}_{s_{i+1},a_{i+1},\cdots}[\sum_{l=0}^\infty\gamma^{l}r(s_{i+l})],
\end{equation}
\begin{equation}\label{V}
    V_{\xi}(s_{i})=\mathbb{E}_{a_{i},s_{i+1},\cdots}[\sum_{l=0}^\infty\gamma^{l}r(s_{i+l})],
\end{equation}
\begin{equation}\label{A}
    A_{\xi}(s_{i},a_i)=Q_{\xi}(s_{i},a_i)-V_{\xi}(s_{i})
\end{equation}
where
\begin{equation}\label{Qas}
    a_{i}\sim P^*_{\xi}(a|s),~~s_{i+1}=F(s_{i},a_{i})
\end{equation}
In addition, define $\nu_{\xi}$ as the discounted visitation frequencies
\begin{equation}\label{nu}
    \nu_{\xi}(s)=p(s_{0}=s)+\gamma p(s_{1}=s)+\gamma^2 p(s_{2}=s)+\cdots
\end{equation}
where $s_{0}\sim p_0$, $a_{i}$ and $s_{i\geq 1}$ are generated according to $P^{*}_{\xi}$ and $F$. Let $\xi^-$ denote the old parameters in last iteration. TRPO proposes to optimize the parameters $\xi$ iteratively regarding the following objective function:

\begin{eqnarray}
    \text{maximize}~&\mathbb{E}_{s\sim\nu_{\xi^-},a\sim q^*}[\frac{P^*_{\xi}(a|s)}{q(a|s)}A_{\xi^-}(a|s)]\label{TRPOobj}\\
    \text{subject to}~&\mathbb{E}_{s\sim\nu_{\xi^-}}[D_{\text{KL}}(P^*_{\xi^-}(\cdot|s)\parallel\label{TRPOconstr} P^*_{\xi}(\cdot|s))]\leq\epsilon
\end{eqnarray}
where $q^*(a|s)$ is the importance sampling distribution and $D_{\text{KL}}(P^*_{\xi^-}\parallel P^{*}_{\xi})$ is the Kullback-Leibler divergence between the old and current policies.

\subsection{Partially Observable TRPO}
As mentioned, our navigation problem is considered as a POMDP. The policy $P_{\theta}(a_i|o_i)$ depends on the observation $o_{i}$ instead of the true state. Therefore, we write the objective function (\ref{TRPOobj}) and the constraint (\ref{TRPOconstr}) as
\begin{eqnarray}
    \text{maximize}~&\mathbb{E}_{s\sim\nu_{\theta^-},a,o\sim q}[\frac{\sum_{o}\beta(o|s)P_{\theta}(a|o)}{q(a,o|s)}A_{\theta^-}(a|s)]\label{POTRPOobj}\\
    \text{subject to}~&\mathbb{E}_{s\sim\nu_{\theta^-}}[D_{\text{KL}}(P_{\theta^-}(\cdot|o)\parallel\label{POTRPOconstr} P_{\theta}(\cdot|o))]\leq\epsilon
\end{eqnarray}
For PO-TRPO, samples are collected by executing the old policy $P_{\theta^-}(a|o)$ to generate a set of trajectories, such as  $s_{0},o_{0},a_{0},s_{1},o_{1},a_{1},\cdots,s_{T-1},o_{T-1},a_{T-1}, s_{T}$. Therefore,
\begin{equation}\label{ISD}
    q(a_i,o_i|s_i)=\beta(o_i|s_i)P_{\theta^{-}}(a_i|o_i)
\end{equation}
where $i=0,\cdots,T-1$.

Next, for a trajectory $s_{0:T}$, we use the generalized advantage estimation (GAE) \cite{schulman2015high} to construct an empirical estimation $\hat{A}$ of the advantage function $A_{\theta^-}(a_i|s_i)$ as the following:
\begin{equation}\label{gae}
    \hat{A}_{i}=\sum_{l=0}^{T-i}(\gamma\lambda)^l\delta_{i+l}^V
\end{equation}
where
\begin{equation}\label{delta}
    \delta_{i}^{V}=r_{i}+\gamma\hat{V}_{\zeta}(s_{i+1})-\hat{V}_{\zeta}(s_i)
\end{equation}
and $\hat{V}_{\zeta}(s_{i})$ is the estimation of the value function (\ref{V}) with parameters $\zeta$ (and $\zeta^-$ being the old parameters). By collecting a set of $K$ trajectories $\{s^{k}_{0:T_{k}},o^{k}_{0:T_{k}},a^{k}_{0:T_{k}}\}_{k=1}^{K}$, $\hat{V}_{\zeta}$ is obtained by solving the following constrained regression problem \cite{schulman2015high}:
\begin{eqnarray}
    \text{minimize}&\hspace{-2mm}J_{1\zeta}=\sum_{k=1}^{K}\hspace{-1mm}\sum_{i=0}^{T_{k}}\hspace{-1mm}\|\hat{V}_{\zeta}(s^{k}_{i})\hspace{-0.5mm}-\hspace{-0.5mm}\sum_{l=0}^{T_{k}-i}\gamma^{l}r^{k}_{i+l}\|^2\label{hatV}\\
    \text{subject to}&\hspace{-2mm}\sum_{k=1}^{K}\sum_{i=0}^{T_{k}}\frac{\|V_{\zeta}(s^{k}_{i})-V_{\zeta^-}(s^{k}_{i})\|}{2J_{1\zeta^-}}\leq\epsilon_1\label{hatVConstr}
\end{eqnarray}

Finally, as the conditional observation probability distribution $\beta(o|s)$ is independent of parameters $\theta$ and time, we obtain an estimation of the objective function (\ref{POTRPOobj}) and the constraints (\ref{POTRPOconstr}) by replacing the expectations with sample averages as:
\begin{eqnarray}
    \text{maximize}&\hspace{-2mm}J_\theta=\frac{1}{\sum_{k=1}^{K}T_{k}}\sum_{k=1}^K\sum_{i=0}^{T_{k}} \frac{P_{\theta}(a^{k}_i|o^{k}_i)}{P_{\theta^{-}}(a^{k}_i|o^{k}_i)}\hat{A}^{k}_i\label{EstPOTRPOobj}\\
    \text{subject to}&\hspace{-2mm}\bar{D}_{\text{KL}}^{\theta^-}(P_{\theta^-},P_{\theta})\leq\epsilon\label{EstPOTRPOconstr}
\end{eqnarray}
where
\begin{equation}\label{barKL}
    \bar{D}_{\text{KL}}^{\theta^-}(P_{\theta^-},P_{\theta})=\frac{1}{\sum_{k=1}^{K}T_{k}}\sum_{k=1}^K\sum_{i=0}^{T_{k}} D_{\text{KL}}(P_{\theta^-}(\cdot|o^k_i)\parallel P_{\theta}(\cdot|o^k_i))]
\end{equation}
which has the same form as the one obtained in \cite{schulman2015high}, except that the policy $P_{\theta}(a|o)$ is conditioned on observation $o$ instead.

Finally, the constrained optimization problem described in (\ref{EstPOTRPOobj}) and (\ref{EstPOTRPOconstr}) is solved by conjugate gradient algorithm \cite{nocedal2006conjugate} algorithm. To summarize, the pseudo code for PO-TRPO update in \textbf{Algorithm} \ref{RPL} is given as below:

\begin{algorithm}[h]
\caption{PO-TRPO}
\begin{algorithmic}\label{PO-TRPO}
\STATE Compute the estimated advantages $\hat{A}_{i}$ for all time steps using GAE with the estimated value function $\hat{V}_{\zeta}$.
\STATE Update $\theta$ with objective function (\ref{EstPOTRPOobj}) and constraints (\ref{EstPOTRPOconstr})
\STATE Update $\zeta$ with objective function (\ref{hatV}) and constraints (\ref{hatVConstr})
\end{algorithmic}
\end{algorithm}

\section{Simulation}
As a data-driven approach, our deep neural network policy requires a massive amount of data to learn the socially concomitant navigation behavior. In this section, we describe how to construct a simulative environment according to the proposed RPL scheme. Particularly, the environments, the deep neural network policy and the PO-TRPO algorithm (\textbf{Algorithm} \ref{PO-TRPO}) are developed under the framework of RLLAB \cite{duan2016benchmarking}. We make use of trajectories of interacting pedestrians collected from five different data sets, which includes the ETH and Hotel video clips from the ETH Walking Pedestrians (EWAP) \cite{pellegrini2009you}, the motion capture (MC) data set from \cite{kretzschmar2016socially}, as well as the Zara and UCY video clips from \cite{lerner2007crowds}. Note that, the Zara and UCY data sets have multiple subsets: Zara01, Zara02, Zara03, UCY01 and UCY03. Thus, there are totally 8 different RPL environments, i.e., $\mathbb{E}=\{\mathcal{E}_1,\cdots,\mathcal{E}_8\}$. The details of these 8 environments are summarized in Tab. \ref{tab.RPL}.
\begin{table}[h]
\caption{Details of the 8 RPL environments}\label{tab.RPL}
\centering
\begin{tabular}{|c|c|c|c|c|}
  \hline
  % after \\: \hline or \cline{col1-col2} \cline{col3-col4} ...
  Name & ETH & Hotel & MC & Zara01 \\
  No. Trajectories & 365 & 420 & 324 & 148 \\
  \hline
  Name & Zara02 & Zara03 & UCY01 & UCY03 \\
  No. Trajectories & 204 & 137 & 413 & 434 \\
  \hline
\end{tabular}
\end{table}

Each trajectory in these environment provides the ID and a sequences of 2-D Cartesian positions of a pedestrian with a sampling period $\Delta t=0.1$ second. In addition, eight binary grid maps $\mathcal{M}_1,\cdots,\mathcal{M}_8$ representing the occupied space/static obstacles are given. However, these maps are kept unknown to the robot throughout training and evaluation. They are only used to simulate the robot's perception to the environment as the state $p_{\text{obs}}$ and observation $\hat{p}_{\text{obs}}$. Without loss of generality, we use the ETH data set as the evaluation environment and all other data sets in the Tab. \ref{tab.RPL} as training environments. In other words, the learned policy's performance will be assessed in an RPL environment that is excluded during training, which reflects whether it can properly generalize to uncovered situations.

As some of the trajectories in these environments are of people who were wandering or remained approximately stationary, they are excluded from the candidates of the robot's companion but will still be considered as pedestrians when the robot is navigating in the same environment.

We use a feed-forward neural network with 2 hidden layers as the feature network in our NN policy, containing 256 and 64 Tanh units, respectively. Its output is then fed to a LSTM network with 64 units. The variance of of the Gaussian output unit $\sigma$ is chosen to be linearly decaying from 0.5 to 0.05 in 100 training iterations, which effectively encourages exploration during the early stage of learning and ensures convergence of the navigation policy. For GAE, a 3 layer feed-forward network with 256,64,16 Tanh units are used, with $\gamma=0.995$ and $\lambda=0.96$. The update step size for policy network is adaptively chosen as $\epsilon=0.01/\sigma$. For GAE update, a fixed step size $\epsilon_1=0.1$ is used. The update batch size (Batch\_size in \textbf{Algorithm} \ref{RPL}) is 50000.

In RPL, we consider at most 3 pedestrians (i.e., $n_{p}=3$). Thus, the state $p_{\text{ped}}$ and observation $\hat{p}_{\text{ped}}$ will only describe the 3 closest pedestrians and omit the others. For the situation where less that 3 pedestrians are perceived, dummy static pedestrians will be created in the remote corner of the environment so as to maintain the dimensions of $p_{\text{ped}}$ and $\hat{p}_{\text{ped}}$.

Considering a Kobuki Turtlebot 2 with a Hokuyo URG-04LX laser range finder \cite{kneip2009characterization} mounted on its top, we specify the sensor limitation of the robot in simulation as follows:
\begin{eqnarray}
% \nonumber to remove numbering (before each equation)
  \phi_{\text{ped}}^+&=&\phi_{\text{obs}}^+=\frac{2\pi}{3}\\
  \phi_{\text{ped}}^-&=&\phi_{\text{obs}}^-=-\frac{2\pi}{3}
\end{eqnarray}

The measurement noises $\tilde{d}_\text{ped},\tilde{d}_\text{com}$ and $\tilde{d}_\text{obs}$ are modeled by zero-mean Gaussian $\mathcal{N}(0,\sigma^2_\text{ped}),\mathcal{N}(0,\sigma^2_\text{com})$ and $\mathcal{N}(0,\sigma^2_\text{obs})$ with their variances specified as follows:
\begin{eqnarray}
% \nonumber to remove numbering (before each equation)
  \sigma_\text{ped}&=&0.01d_{\text{ped}}^j \label{sim_noise_ped}\\
  \sigma_\text{com} &=& 0.01d_{\text{com}}^j \label{sim_noise_com}\\
  \sigma_\text{obs} &=& 0.01d_{\text{obs}}^j\label{sim_noise_obs}
\end{eqnarray}

\begin{figure}[h]\centering
\subfigure[Real-world Environment]{\label{fig.content-driven.1}\includegraphics[width=0.48\hsize]{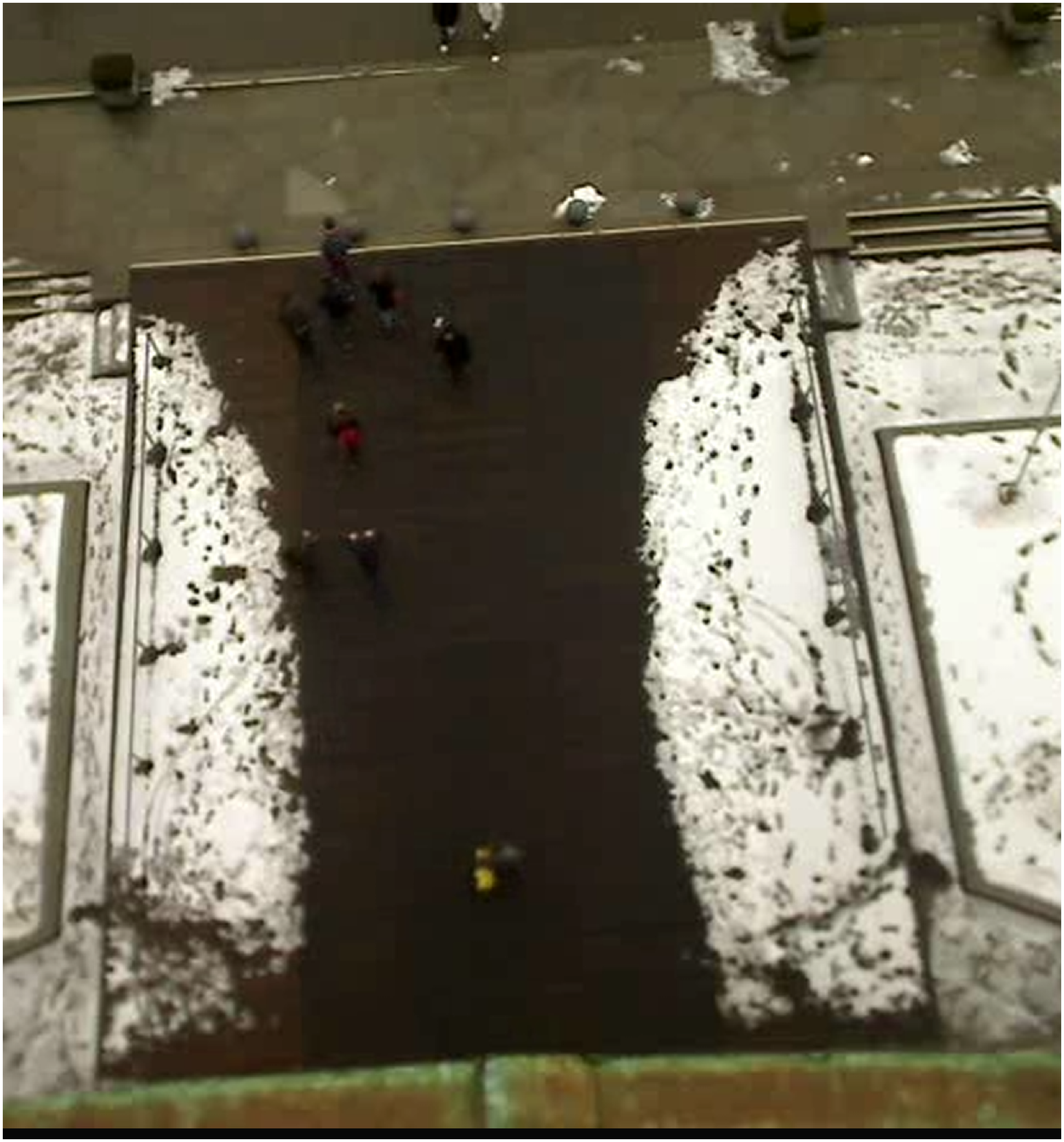}}
\subfigure[Simulative Environment for RPL]{\label{fig.content-driven.2}\includegraphics[width=0.48\hsize]{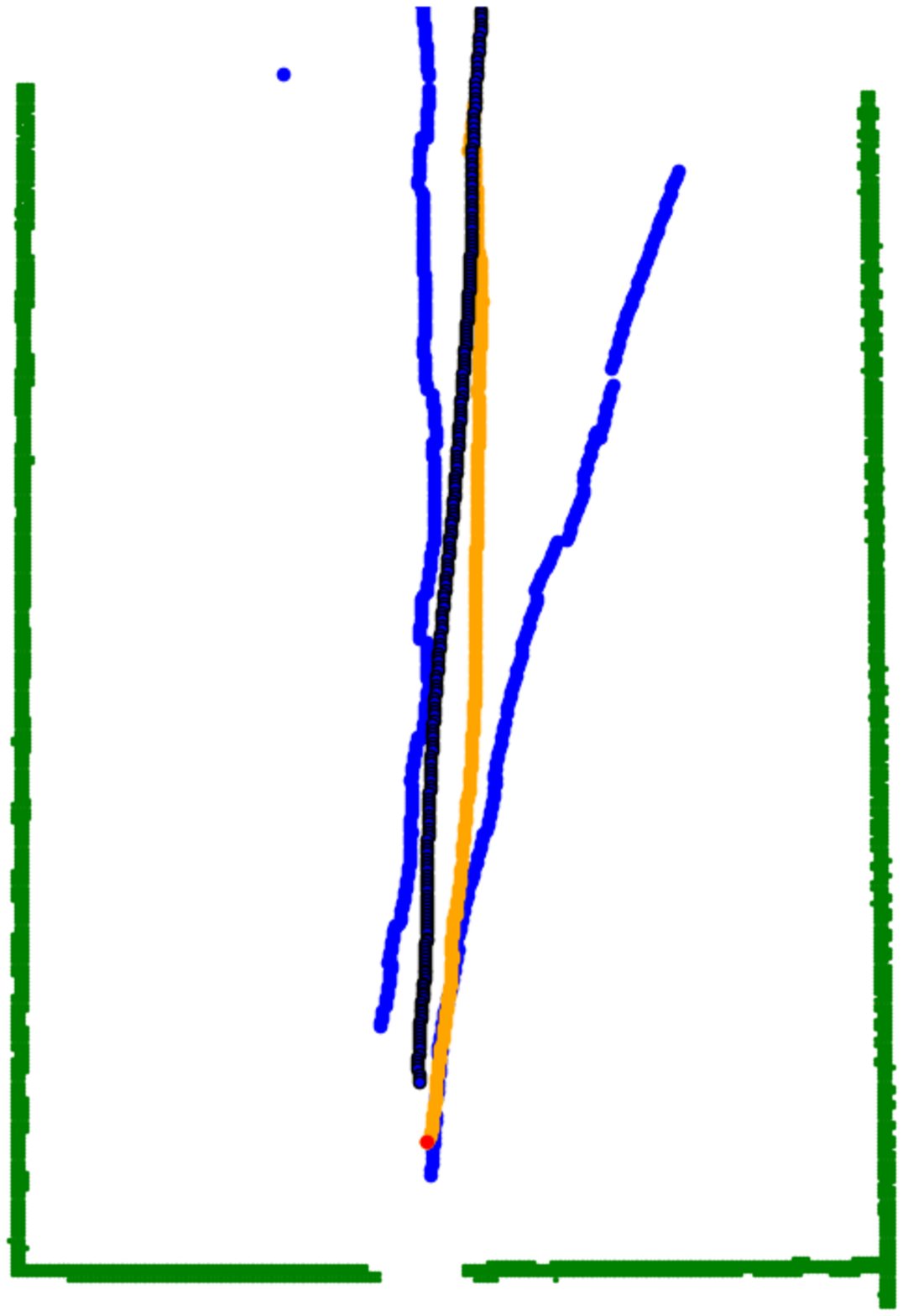}}
\caption{An illustrative example our RPL simulative environment. The black curve represents the trajectory of the robot navigating toward its goal (the red dot). The yellow curve denotes the trajectory of the robot's companion. Besides, there are a number of blue curves representing the pedestrians perceived by the robot and the green lines denotes the fences around the entrance of the university (bottom center). Note that, all trajectories of pedestrians are not synthesized but captured from the video. Thus, the robot can be thought as playing a role as an extra person in an realistic environment.}\label{fig.sim_exp}
\end{figure}

Finally, the maximum translational and rotational velocities are assigned as $0.7\text{m/s}$ and $\frac{\pi}{3}\text{rad/s}$, i.e., $0\leq v_{T}\leq 0.7$ and $|v_R|\leq\frac{\pi}{3}$ and $\Delta t=0.1$. An example of our RPL environment constructed from the ETH data set is illustrated in Fig. \ref{fig.sim_exp}

\subsection{Results}
\begin{figure}[h]
  \centering
  % Requires \usepackage{graphicx}
  \includegraphics[width=\hsize]{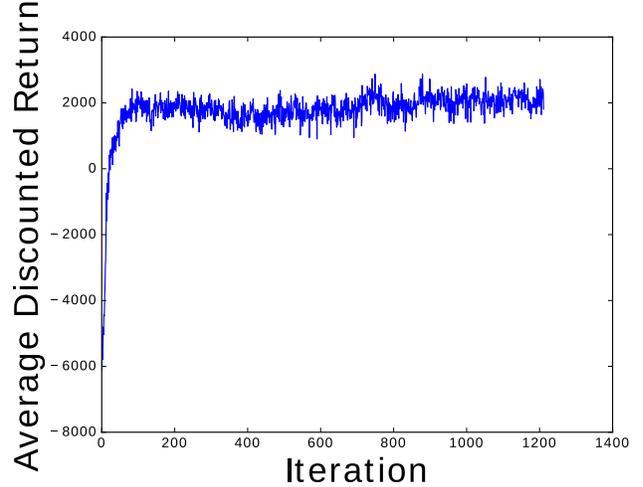}\\
  \caption{Average discounted return as RPL progresses}\label{fig.sim.adr}
\end{figure}

We trained our deep policy network for 1200 iterations with the data from RPL environments except for the held-out ETH environment. The curve of average discounted return obtained from each batch of trajectories is visualized in Fig. \ref{fig.sim.adr}

We compare the performance of our policy with a planner based on RVO \cite{van2011reciprocal}, where the robot, its companion and the surrounding pedestrians are treated agents. In every time steps, the positions and velocities of all agents are given to the planner. Note that, for fair comparison, the agents' positions are subject to noise described in (\ref{sim_noise_ped}) and (\ref{sim_noise_com}). For observations to obstacles, we assume the planner has full and perfect knowledge as required in the original RVO algorithm. With this protocol, we update the robot's position according to the planner's output and update the positions of the other agents according to their own trajectories in the RPL environments. The same termination conditions in Section III are applied to the robot directed by the RVO-based planner to determine whether the robot has conducted an successful navigation. For both of our policy and the RVO-based planner, we conduct 300 trials in the evaluation environment and compute the rates (in percentages) of different terminal conditions (RG: the robot reaches the goal successfully; HC/HP/HO: the robot hits a companion/pedestrian/obstacle; and LC: the robot loses its companion). The performance statistics of our policy and the RVO-based planner in SCN scenarios are listed in Tab. \ref{tab.sim.stats}.

\begin{table}[h]
\caption{Rates of different terminal conditions of our policy and RVO-based planner in SCN scenarios}\label{tab.sim.stats}
\centering
\begin{tabular}{|c|c|c|c|c|c|c|c|}
  \hline
  % after \\: \hline or \cline{col1-col2} \cline{col3-col4} ...
  Terminal Condition & RG & LC & HC & HP & HO\\
  \hline
  Our policy & 77 & 5.7 & 6.7 & 9.6 & 1\\
  \hline
  RVO & 29.7 & 47 & 0.3 & 23 & 0 \\
  \hline
\end{tabular}
\end{table}
It can bee seen from Tab. \ref{tab.sim.stats} that our policy performs much better than the RVO-based planner in SCN. The RVO-based planner has a much lower success rate (29.7\%) while its rate of LC is $47\%$, suggesting that it frequently losses its companion in SCN. Clearly, this is due to the fact that RVO is in nature a collision avoidance algorithm. Thus, it simply takes the robot's companion as another normal agent and the robot tends to stay far behind its companion to avoid collision instead of actively following it. On the contrary, our policy achieves a much higher success rate (77\%). This indicates that it learns to effectively balance the objectives of SCN so that the robot is able to reach the prescribed goal while maintaining its distance to its companion and avoiding collision with other agents in the environment.

In addition to SCN, the scenarios without companion are also tested, which, as analyzed in the previous sections, reduces to the traditional social navigation scenarios. The comparative results are shown in Tab. \ref{tab.sim.nonCom.stats}.
\begin{table}[h]
\caption{Rates of different terminal conditions of our policy and RVO-based planner in traditional social navigation scenarios }\label{tab.sim.nonCom.stats}
\centering
\begin{tabular}{|c|c|c|c|c|c|}
  \hline
  % after \\: \hline or \cline{col1-col2} \cline{col3-col4} ...
  Terminal Condition & RG & HP & HO\\
  \hline
  Our policy & 84 & 13.7 & 2.3\\
  \hline
  RVO & 80 & 18 & 2\\
  \hline
\end{tabular}
\end{table}

For situations without companion, our policy still outperforms the RVO-based planner with higher success rate (84\% to 80\%) and lower HP rate (13.7\% to 18\%).

Finally, it is worth noting that the RVO-based planner requires velocities of the companion/pedestrians and an accurate global map of the static obstacles. Conversely, our policy depends only on position measurements that are directly accessible from the robot's onboard sensors, which is therefore much simpler and more practical.
%
%In addition to the above performance metrics, a Turing test is conducted to evaluate whether our policy and the RVO-based planner are able to generate human-like trajectory in social navigation. Specifically, the scenario without companion is considered such that the human trajectory observed in the same scenario can serve as the ground truth. 3 sets of 10 trajectories from our policy, the RVO-based planner and the human ground truth, respectively are collected. The animations of these 30 trajectories are presented to a group of 10 human subjects, who have no knowledge of the source of these trajectories. They are asked to determine whether a presented trajectory is from human demonstration or generated by machines (i.e., our policy or the RVO-based planner). The rates of being perceived as a human-generated trajectory are computed for the 3 sets of data and results are summarized in Fig.

\section{Experiments}
In experiments, we assess the performance of our developed navigation policy by comparing it with humans in the same scenarios. Particularly, a robot and a human are to repeat each specific navigation scenario for 10 times, respectively. Then, the the following two metrics are calculated:

\begin{enumerate}
  \item Average minimum distance to the pedestrians ($\bar{D}_{\text{ped}}$): the average of the minimum distance between the robot/compared human to other pedestrians throughout a trajectory .
  \item Average maximum distance to the companion ($\bar{D}_{\text{com}}$): the average of the maximum distance between the robot/compared human to its/his companion throughout a trajectory.
\end{enumerate}

\begin{figure}[h]
  \centering
  % Requires \usepackage{graphicx}
  \includegraphics[width=0.4\hsize]{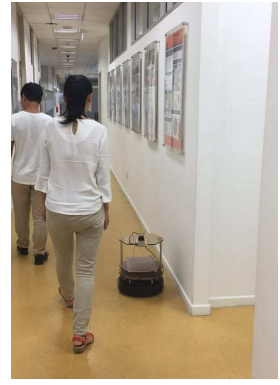}\\
  \caption{The narrow corridor where experiments are performed}\label{fig.exp.corridor}
\end{figure}

We use the same mobile platform (a synchron-drive Turtlebot 2 with a Kobuki base) and the same laser range finder (Hokuyo URG-04LX) simulated in last section. For pedestrian detection and localization, we adopt the ROS-compatible leg tracker in \cite{leigh2015person}. We use an ultra wideband (UWB) indoor positioning system to localize the companion and the navigation goal, which can then be easily mapped to the observations $\hat{p}_\text{com}$ and $d_g,\phi_g$ based on the odometry of the robot. Finally, a laptop is placed onboard as the processing unit and the policy is operated with a period of 0.1 second. The experiments are conducted in a narrow corridor with width of 1.56 meters as shown in Fig. \ref{fig.exp.corridor}, which is a typical scenario that requires pedestrians to navigate cooperatively.

\subsection{Scenario 1: Traditional Social Navigation}
\begin{figure}
  \centering
  % Requires \usepackage{graphicx}
  \subfigure[Trajectories of the robot (moving from left to right) and other two pedestrians (moving from right to left). ]{\label{fig.noComp.robot}\includegraphics[width=\hsize]{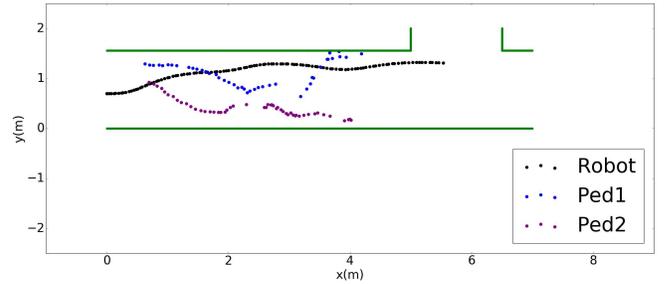}}
  \subfigure[Human control experiment in a similar navigation scenario. The black trajectory is from left to right and the other two are from right to left.]{\label{fig.noComp.human}\includegraphics[width=\hsize]{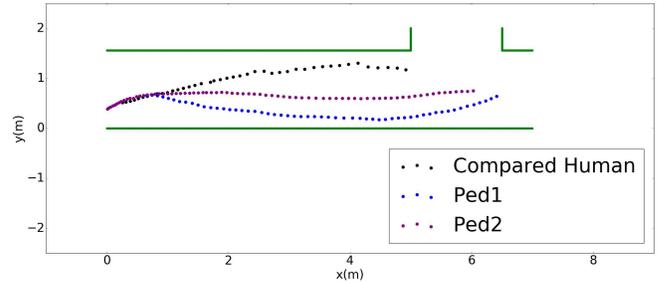}}
  \caption{Comparison between the robot with our policy and human control experiment in a social navigation scenario}\label{fig.noComp}
\end{figure}

In this subsection, we examine our method's performance in traditional social navigation scenario. Particularly, the robot is required to pass the corridor with two oncoming pedestrians and arrive at a goal that is 7 meters ahead. In addition, a control experiment of 3 humans (one as the compared human and the other two as pedestrians) is conducted in the same space. The metric $\bar{D}_{\text{ped}}$ is computed. Example trajectories of the robot and the human control are shown in Fig. \ref{fig.noComp}. In the robotic experiments, the trajectories of pedestrians are obtained from the robot's laser range finder while the robot's trajectory is based on its own odometry sensor. On the other hand, all trajectories in the human control experiments are captured using the UWB localization system.

From Fig. \ref{fig.noComp}, it is clear that the robot with our policy is able understand human's cooperative behavior for collision avoidance and navigate in an appropriate manner such that both itself and the other two pedestrians can successfully pass through the corridor. Specifically, when observing the two pedestrians (blue and purple) 4 meters ahead. The robot started to approach the wall on its left side so as to create free space on the right for the pedestrians to smoothly walk through. By comparing both figures in Fig. \ref{fig.noComp}, we can see that the robot is as proactive as human since both black trajectories in Fig. \ref{fig.noComp.robot} and Fig. \ref{fig.noComp.human} started to make space for the oncoming pedestrians at the early stage of cooperative avoidance process. As for the performance metrics, the average minimal distance to pedestrians for our robot is $\bar{D}_{\text{ped}}=0.35$m. Although it is smaller than that of the human control experiments ($\bar{D}_{\text{ped}}$=0.56m), this value still indicates a safe and decent navigation behavior of our robot as its radius is only 0.17m.

\subsection{Scenario 2: Socially Concomitant Navigation}
\begin{figure}
  \centering
  % Requires \usepackage{graphicx}
  \subfigure[Trajectories of the robot and its companion (moving from left to right) and a pedestrian (moving from right to left). ]{\label{fig.content-driven.1}\includegraphics[width=\hsize]{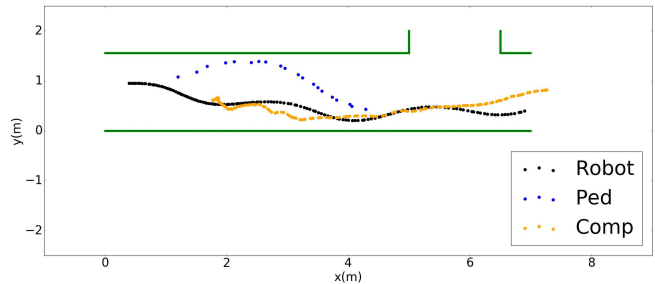}}
  \subfigure[Human control experiment in a similar SCN. The black (compared human) and orange (companion) trajectories are from left to right and the blue (pedestrian) trajectory is from right to left.]{\label{fig.content-driven.2}\includegraphics[width=\hsize]{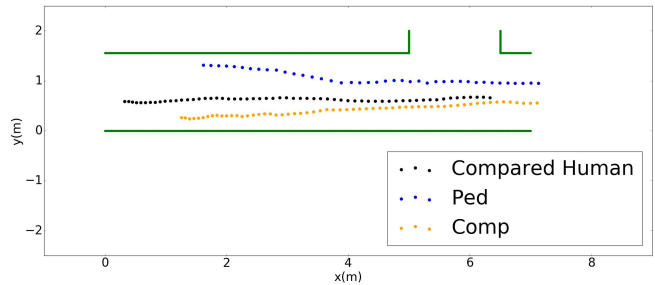}}
  \caption{Comparison between the robot with our policy and human control experiment in a SCN scenario}\label{fig.SCN}
\end{figure}

In this subsection, the scenario of SCN is studied. A human companion initially standing in front of the robot will start to walk through the same corridor while another pedestrian is passing from the other end. As described in the previous sections, the robot with our policy should closely navigate with its companion and avoid the oncoming pedestrian cooperatively. An additional metric $\bar{D}_{\text{com}}$ is used to evaluate the performance of our policy by comparing with the statistics obtained from another 10 human control experiments. Example trajectories are shown in Fig. \ref{fig.SCN} and the performance metrics $\bar{D}_{\text{ped}}$ and $\bar{D}_{\text{com}}$ are summarized in Tab. \ref{tab.exp.scn.stats}.

\begin{table}[h]
\caption{Performance metrics of the robot and human controls in SCN scenarios}\label{tab.exp.scn.stats}
\centering
\begin{tabular}{|c|c|c|}
  \hline
  % after \\: \hline or \cline{col1-col2} \cline{col3-col4} ...
   & $\bar{D}_{\text{ped}}$(m) & $\bar{D}_{\text{com}}$(m)\\
  \hline
  Robot & 0.49 & 1.05 \\
  \hline
  Compared Human & 0.37 & 1\\
  \hline
\end{tabular}
\end{table}

As shown in Fig. \ref{fig.SCN} and Tab. \ref{tab.exp.scn.stats}, the robot is able to achieve both objectives of SCN. On one hand, it is effectively engaged into the joint collision avoidance process. The resulted behavior is similar to that observed in the last subsection and the robot even has a slightly larger $\bar{D}_{\text{ped}}$. On the other hand, the average maximum distance $\bar{D}_{\text{com}}$ is 1.05m, which is within the limit (2m) we specified in the learning process and nearly the same as that of the compared human, showing that the robot can actively navigate along with its companion instead of deviating to other areas or lagging itself behind. This shows that the robot driven by our policy is able to understand the pace of its companion and achieve a similar sense of companionship in terms of distance.

In sum, the above results demonstrate the practical efficacy of our methods for both the traditional social navigation and the more complicated SCN scenarios. It proves that the policy learned from our RPL simulative environment is transferable to uncovered real-world situations.

\section{Conclusions}
In this paper, the problem of socially concomitant navigation (SCN) has been investigated and formulated under a POMDP framework, with explicit considerations of the limitation and inaccuracy of mobile robots' onboard sensors. The Partially Observable TRPO (PO-TRPO) algorithm has been proposed for optimization of navigation policies. The Role Playing Learning (RPL) scheme has been developed to enable efficient and safe reinforcement learning of navigation policies by mirroring a large amount of real-world pedestrian trajectories into simulative environments. Comparative simulation and experiment studies have demonstrated the efficacy and superiority of our  policy in both SCN and traditional social navigation scenarios.

\bibliographystyle{IEEEtran}

\begin{thebibliography}{10}
\providecommand{\url}[1]{#1}
\csname url@samestyle\endcsname
\providecommand{\newblock}{\relax}
\providecommand{\bibinfo}[2]{#2}
\providecommand{\BIBentrySTDinterwordspacing}{\spaceskip=0pt\relax}
\providecommand{\BIBentryALTinterwordstretchfactor}{4}
\providecommand{\BIBentryALTinterwordspacing}{\spaceskip=\fontdimen2\font plus
\BIBentryALTinterwordstretchfactor\fontdimen3\font minus
  \fontdimen4\font\relax}
\providecommand{\BIBforeignlanguage}[2]{{%
\expandafter\ifx\csname l@#1\endcsname\relax
\typeout{** WARNING: IEEEtran.bst: No hyphenation pattern has been}%
\typeout{** loaded for the language `#1'. Using the pattern for}%
\typeout{** the default language instead.}%
\else
\language=\csname l@#1\endcsname
\fi
#2}}
\providecommand{\BIBdecl}{\relax}
\BIBdecl

\bibitem{thrun1997dynamic}
D.~F. W. B.~S. Thrun, D.~Fox, and W.~Burgard, ``The dynamic window approach to
  collision avoidance,'' \emph{IEEE Transactions on Robotics and Automation},
  vol.~4, p.~1, 1997.

\bibitem{hwang1992potential}
Y.~K. Hwang and N.~Ahuja, ``A potential field approach to path planning,''
  \emph{IEEE Transactions on Robotics and Automation}, vol.~8, no.~1, pp.
  23--32, 1992.

\bibitem{ge2000new}
S.~S. Ge and Y.~J. Cui, ``New potential functions for mobile robot path
  planning,'' \emph{IEEE Transactions on robotics and automation}, vol.~16,
  no.~5, pp. 615--620, 2000.

\bibitem{trautman2010unfreezing}
P.~Trautman and A.~Krause, ``Unfreezing the robot: Navigation in dense,
  interacting crowds,'' in \emph{Intelligent Robots and Systems (IROS), 2010
  IEEE/RSJ International Conference on}, 2010, pp. 797--803.

\bibitem{trautman2015robot}
P.~Trautman, J.~Ma, R.~M. Murray, and A.~Krause, ``Robot navigation in dense
  human crowds: Statistical models and experimental studies of human--robot
  cooperation,'' \emph{The International Journal of Robotics Research},
  vol.~34, no.~3, pp. 335--356, 2015.

\bibitem{helbing1995social}
D.~Helbing and P.~Molnar, ``Social force model for pedestrian dynamics,''
  \emph{Physical review E}, vol.~51, no.~5, p. 4282, 1995.

\bibitem{helbing2000simulating}
D.~Helbing, I.~Farkas, and T.~Vicsek, ``Simulating dynamical features of escape
  panic,'' \emph{Nature}, vol. 407, no. 6803, pp. 487--490, 2000.

\bibitem{van2011reciprocal}
J.~Van Den~Berg, S.~J. Guy, M.~Lin, and D.~Manocha, ``Reciprocal n-body
  collision avoidance,'' in \emph{Robotics research}, 2011, pp. 3--19.

\bibitem{van2011lqg}
J.~Van Den~Berg, P.~Abbeel, and K.~Goldberg, ``Lqg-mp: Optimized path planning
  for robots with motion uncertainty and imperfect state information,''
  \emph{The International Journal of Robotics Research}, vol.~30, no.~7, pp.
  895--913, 2011.

\bibitem{van2008reciprocal}
J.~Van~den Berg, M.~Lin, and D.~Manocha, ``Reciprocal velocity obstacles for
  real-time multi-agent navigation,'' in \emph{Robotics and Automation, 2008.
  ICRA 2008. IEEE International Conference on}, 2008, pp. 1928--1935.

\bibitem{pellegrini2009you}
S.~Pellegrini, A.~Ess, K.~Schindler, and L.~Van~Gool, ``You'll never walk
  alone: Modeling social behavior for multi-target tracking,'' in \emph{2009
  IEEE 12th International Conference on Computer Vision}, 2009, pp. 261--268.

\bibitem{yamaguchi2011you}
K.~Yamaguchi, A.~C. Berg, L.~E. Ortiz, and T.~L. Berg, ``Who are you with and
  where are you going?'' in \emph{Computer Vision and Pattern Recognition
  (CVPR), 2011 IEEE Conference on}, 2011, pp. 1345--1352.

\bibitem{kuderer2012feature}
M.~Kuderer, H.~Kretzschmar, C.~Sprunk, and W.~Burgard, ``Feature-based
  prediction of trajectories for socially compliant navigation.'' in
  \emph{Robotics: science and systems}, 2012.

\bibitem{kretzschmar2016socially}
H.~Kretzschmar, M.~Spies, C.~Sprunk, and W.~Burgard, ``Socially compliant
  mobile robot navigation via inverse reinforcement learning,'' \emph{The
  International Journal of Robotics Research}, p. 0278364915619772, 2016.

\bibitem{kim2016socially}
B.~Kim and J.~Pineau, ``Socially adaptive path planning in human environments
  using inverse reinforcement learning,'' \emph{International Journal of Social
  Robotics}, vol.~8, no.~1, pp. 51--66, 2016.

\bibitem{bicchi2010towards}
A.~Bicchi, A.~Fagiolini, and L.~Pallottino, ``Towards a society of robots,''
  \emph{IEEE Robotics \& Automation Magazine}, vol.~17, no.~4, pp. 26--36,
  2010.

\bibitem{gross2011progress}
H.-M. Gross, C.~Schroeter, S.~Mueller, M.~Volkhardt, E.~Einhorn, A.~Bley,
  C.~Martin, T.~Langner, and M.~Merten, ``Progress in developing a socially
  assistive mobile home robot companion for the elderly with mild cognitive
  impairment,'' in \emph{Intelligent Robots and Systems (IROS), 2011 IEEE/RSJ
  International Conference on}, 2011, pp. 2430--2437.

\bibitem{wang2014adaptive}
H.~Wang and X.~P. Liu, ``Adaptive shared control for a novel mobile assistive
  robot,'' \emph{IEEE/ASME Transactions on Mechatronics}, vol.~19, no.~6, pp.
  1725--1736, 2014.

\bibitem{argall2009survey}
B.~D. Argall, S.~Chernova, M.~Veloso, and B.~Browning, ``A survey of robot
  learning from demonstration,'' \emph{Robotics and autonomous systems},
  vol.~57, no.~5, pp. 469--483, 2009.

\bibitem{abbeel2004apprenticeship}
P.~Abbeel and A.~Y. Ng, ``Apprenticeship learning via inverse reinforcement
  learning,'' in \emph{Proceedings of the twenty-first international conference
  on Machine learning}, 2004, p.~1.

\bibitem{ziebart2008maximum}
B.~D. Ziebart, A.~L. Maas, J.~A. Bagnell, and A.~K. Dey, ``Maximum entropy
  inverse reinforcement learning.'' in \emph{AAAI}, vol.~8, 2008, pp.
  1433--1438.

\bibitem{ratliff2006maximum}
N.~D. Ratliff, J.~A. Bagnell, and M.~A. Zinkevich, ``Maximum margin planning,''
  in \emph{Proceedings of the 23rd international conference on Machine
  learning}, 2006, pp. 729--736.

\bibitem{ziebart2009planning}
B.~D. Ziebart, N.~Ratliff, G.~Gallagher, C.~Mertz, K.~Peterson, J.~A. Bagnell,
  M.~Hebert, A.~K. Dey, and S.~Srinivasa, ``Planning-based prediction for
  pedestrians,'' in \emph{Intelligent Robots and Systems, 2009. IROS 2009.
  IEEE/RSJ International Conference on}, 2009, pp. 3931--3936.

\bibitem{henry2010learning}
P.~Henry, C.~Vollmer, B.~Ferris, and D.~Fox, ``Learning to navigate through
  crowded environments,'' in \emph{Robotics and Automation (ICRA), 2010 IEEE
  International Conference on}, 2010, pp. 981--986.

\bibitem{vernaza2012efficient}
P.~Vernaza and D.~Bagnell, ``Efficient high dimensional maximum entropy
  modeling via symmetric partition functions,'' in \emph{Advances in Neural
  Information Processing Systems}, 2012, pp. 575--583.

\bibitem{kitani2012activity}
K.~M. Kitani, B.~D. Ziebart, J.~A. Bagnell, and M.~Hebert, ``Activity
  forecasting,'' in \emph{European Conference on Computer Vision}.\hskip 1em
  plus 0.5em minus 0.4em\relax Springer, 2012, pp. 201--214.

\bibitem{choi2011map}
J.~Choi and K.-E. Kim, ``Map inference for bayesian inverse reinforcement
  learning,'' in \emph{Advances in Neural Information Processing Systems},
  2011, pp. 1989--1997.

\bibitem{kim2011gaussian}
K.~Kim, D.~Lee, and I.~Essa, ``Gaussian process regression flow for analysis of
  motion trajectories,'' in \emph{Computer vision (ICCV), 2011 IEEE
  international conference on}, 2011, pp. 1164--1171.

\bibitem{alahi2016social}
A.~Alahi, K.~Goel, V.~Ramanathan, A.~Robicquet, L.~Fei-Fei, and S.~Savarese,
  ``Social lstm: Human trajectory prediction in crowded spaces,'' in
  \emph{Proceedings of the IEEE Conference on Computer Vision and Pattern
  Recognition}, 2016, pp. 961--971.

\bibitem{robicquet2016learning}
A.~Robicquet, A.~Sadeghian, A.~Alahi, and S.~Savarese, ``Learning social
  etiquette: Human trajectory understanding in crowded scenes,'' in
  \emph{European Conference on Computer Vision}, 2016, pp. 549--565.

\bibitem{johansson2007specification}
A.~Johansson, D.~Helbing, and P.~K. Shukla, ``Specification of the social force
  pedestrian model by evolutionary adjustment to video tracking data,''
  \emph{Advances in complex systems}, vol.~10, no. supp02, pp. 271--288, 2007.

\bibitem{lerner2007crowds}
A.~Lerner, Y.~Chrysanthou, and D.~Lischinski, ``Crowds by example,'' in
  \emph{Computer Graphics Forum}, vol.~26, no.~3.\hskip 1em plus 0.5em minus
  0.4em\relax Wiley Online Library, 2007, pp. 655--664.

\bibitem{helbing2011pedestrian}
D.~Helbing and A.~Johansson, ``Pedestrian, crowd and evacuation dynamics,'' in
  \emph{Encyclopedia of Complexity and Systems Science}, 2009, pp. 6476--6495.

\bibitem{muller2008socially}
J.~M{\"u}ller, C.~Stachniss, K.~Arras, and W.~Burgard, ``Socially inspired
  motion planning for mobile robots in populated environments,'' in \emph{Proc.
  of International Conference on Cognitive Systems}, 2008.

\bibitem{mehta2016autonomous}
D.~Mehta, G.~Ferrer, and E.~Olson, ``Autonomous navigation in dynamic social
  environments using multi-policy decision making,'' in \emph{Intelligent
  Robots and Systems (IROS), 2016 IEEE/RSJ International Conference on}, 2016,
  pp. 1190--1197.

\bibitem{foka2010probabilistic}
A.~F. Foka and P.~E. Trahanias, ``Probabilistic autonomous robot navigation in
  dynamic environments with human motion prediction,'' \emph{International
  Journal of Social Robotics}, vol.~2, no.~1, pp. 79--94, 2010.

\bibitem{seder2007dynamic}
M.~Seder and I.~Petrovic, ``Dynamic window based approach to mobile robot
  motion control in the presence of moving obstacles,'' in \emph{Robotics and
  Automation, 2007 IEEE International Conference on}, 2007, pp. 1986--1991.

\bibitem{fiorini1998motion}
P.~Fiorini and Z.~Shiller, ``Motion planning in dynamic environments using
  velocity obstacles,'' \emph{The International Journal of Robotics Research},
  vol.~17, no.~7, pp. 760--772, 1998.

\bibitem{schulman2015high}
J.~Schulman, P.~Moritz, S.~Levine, M.~Jordan, and P.~Abbeel, ``High-dimensional
  continuous control using generalized advantage estimation,'' \emph{arXiv
  preprint arXiv:1506.02438}, 2015.

\bibitem{schulman2015trust}
J.~Schulman, S.~Levine, P.~Moritz, M.~I. Jordan, and P.~Abbeel, ``Trust region
  policy optimization,'' \emph{CoRR, abs/1502.05477}, 2015.

\bibitem{lecun2015deep}
Y.~LeCun, Y.~Bengio, and G.~Hinton, ``Deep learning,'' \emph{Nature}, vol. 521,
  no. 7553, pp. 436--444, 2015.

\bibitem{pfeiffer2016perception}
M.~Pfeiffer, M.~Schaeuble, J.~Nieto, R.~Siegwart, and C.~Cadena, ``From
  perception to decision: A data-driven approach to end-to-end motion planning
  for autonomous ground robots,'' \emph{arXiv preprint arXiv:1609.07910}, 2016.

\bibitem{chen2016decentralized}
Y.~F. Chen, M.~Liu, M.~Everett, and J.~P. How, ``Decentralized
  non-communicating multiagent collision avoidance with deep reinforcement
  learning,'' \emph{arXiv preprint arXiv:1609.07845}, 2016.

\bibitem{zhu2016target}
Y.~Zhu, R.~Mottaghi, E.~Kolve, J.~J. Lim, A.~Gupta, L.~Fei-Fei, and A.~Farhadi,
  ``Target-driven visual navigation in indoor scenes using deep reinforcement
  learning,'' \emph{arXiv preprint arXiv:1609.05143}, 2016.

\bibitem{choi2016extrinsic}
D.-G. Choi, Y.~Bok, J.-S. Kim, and I.~S. Kweon, ``Extrinsic calibration of 2-d
  lidars using two orthogonal planes,'' \emph{IEEE Transactions on Robotics},
  vol.~32, no.~1, pp. 83--98, 2016.

\bibitem{miller2015optimal}
L.~M. Miller and T.~D. Murphey, ``Optimal planning for target localization and
  coverage using range sensing,'' in \emph{Automation Science and Engineering
  (CASE), 2015 IEEE International Conference on}, 2015, pp. 501--508.

\bibitem{endres20143}
F.~Endres, J.~Hess, J.~Sturm, D.~Cremers, and W.~Burgard, ``3-d mapping with an
  rgb-d camera,'' \emph{IEEE Transactions on Robotics}, vol.~30, no.~1, pp.
  177--187, 2014.

\bibitem{foix2010object}
S.~Foix, G.~Alenya, J.~Andrade-Cetto, and C.~Torras, ``Object modeling using a
  tof camera under an uncertainty reduction approach,'' in \emph{Robotics and
  Automation (ICRA), 2010 IEEE International Conference on}, 2010, pp.
  1306--1312.

\bibitem{liu2014topological}
M.~Liu and R.~Siegwart, ``Topological mapping and scene recognition with
  lightweight color descriptors for an omnidirectional camera,'' \emph{IEEE
  Transactions on Robotics}, vol.~30, no.~2, pp. 310--324, 2014.

\bibitem{kneip2009characterization}
L.~Kneip, F.~T{\^a}che, G.~Caprari, and R.~Siegwart, ``Characterization of the
  compact hokuyo urg-04lx 2d laser range scanner,'' in \emph{Robotics and
  Automation, 2009. ICRA'09. IEEE International Conference on}, 2009, pp.
  1447--1454.

\bibitem{hochreiter1997long}
S.~Hochreiter and J.~Schmidhuber, ``Long short-term memory,'' \emph{Neural
  computation}, vol.~9, no.~8, pp. 1735--1780, 1997.

\bibitem{nocedal2006conjugate}
J.~Nocedal and S.~J. Wright, ``Conjugate gradient methods,'' \emph{Numerical
  optimization}, pp. 101--134, 2006.

\bibitem{duan2016benchmarking}
Y.~Duan, X.~Chen, R.~Houthooft, J.~Schulman, and P.~Abbeel, ``Benchmarking deep
  reinforcement learning for continuous control,'' \emph{arXiv preprint
  arXiv:1604.06778}, 2016.

\bibitem{leigh2015person}
A.~Leigh, J.~Pineau, N.~Olmedo, and H.~Zhang, ``Person tracking and following
  with 2d laser scanners,'' in \emph{Robotics and Automation (ICRA), 2015 IEEE
  International Conference on}, 2015, pp. 726--733.

\end{thebibliography}

\end{document}